\documentclass[journal]{IEEEtran}

\usepackage{xcolor,soul,framed} 

\colorlet{shadecolor}{yellow}
\usepackage[pdftex]{graphicx}
\graphicspath{{../pdf/}{../jpeg/}}
\DeclareGraphicsExtensions{.pdf,.jpeg,.png}

\usepackage{hyperref}  

\begin{document}
\title{Masked conditional variational autoencoders for chromosome straightening}

\author{Jingxiong Li$^1$, Sunyi Zheng$^1$, Zhongyi Shui, Shichuan Zhang, Linyi Yang, Yuxuan Sun, Yunlong Zhang, Honglin Li, Yuanxin Ye, Peter M.A. van Ooijen, Kang Li, Lin Yang* 
\thanks{ J. Li and S. Zheng contributed equally to this work. (corresponding author: Lin Yang)
}
\thanks{J. Li, Z. Shui, S. Zhang, L.Yang, Y. Sun, Y. Zhang, and H. Li, are with College of Computer Science and Technology, Zhejiang University, Hangzhou, 310027, China, School of Engineering, Westlake University, Hangzhou, 310030, China and Institute of Advanced Technology, Westlake Institute for Advanced Study, Hangzhou, 310024, China e-mail: \{lijingxiong, shuizhongyi, zhangshichuan, yanglinyi, sunyuxuan, zhangyunlong, lihonglin\}@westlake.edu.cn).}
\thanks{S. Zheng and L. Yang are with School of Engineering, Westlake University, Hangzhou, 310030, China, Institute of Advanced Technology, Westlake Institute for Advanced Study, Hangzhou, 310024, China (e-mail : \{zhengsunyi, yanglin\}@westlake.edu.cn)}
\thanks{Y. Ye and K. Li are with Department of Laboratory Medicine, West China hospital, Sichuan University, Chengdu, 610000, China (e-mail:\ jennyyyx1982@126.com, \ likang@wchscu.cn).}
\thanks{P.M.A.Ooijen is with University Medical Center Groningen, University of Groningen, Groningen, 9713 GZ, The Netherlands (e-mail:\ p.m.a.van.ooijen@umcg.nl)}}

\maketitle

\begin{abstract}
Karyotyping is of importance for detecting chromosomal aberrations in human disease. However, chromosomes easily appear curved in microscopic images, which prevents cytogeneticists from analyzing chromosome types. To address this issue, we propose a framework for chromosome straightening, which comprises a preliminary processing algorithm and a generative model called masked conditional variational autoencoders (MC-VAE). The processing method utilizes patch rearrangement to address the difficulty in erasing low degrees of curvature, providing reasonable preliminary results for the MC-VAE. The MC-VAE further straightens the results by leveraging chromosome patches conditioned on their curvatures to learn the mapping between banding patterns and conditions. During model training, we apply a masking strategy with a high masking ratio to train the MC-VAE with eliminated redundancy. This yields a non-trivial reconstruction task, allowing the model to effectively preserve chromosome banding patterns and structure details in the reconstructed results. Extensive experiments on three public datasets with two stain styles show that our framework surpasses the performance of state-of-the-art methods in retaining banding patterns and structure details. Compared to using real-world bent chromosomes, the use of high-quality straightened chromosomes generated by our proposed method can improve the performance of various deep learning models for chromosome classification by a large margin. Such a straightening approach has the potential to be combined with other karyotyping systems to assist cytogeneticists in chromosome analysis.
\end{abstract}


\section{Introduction}
\label{sec:introduction}
\begin{figure}[!t]
\centering
\includegraphics[scale=0.40]{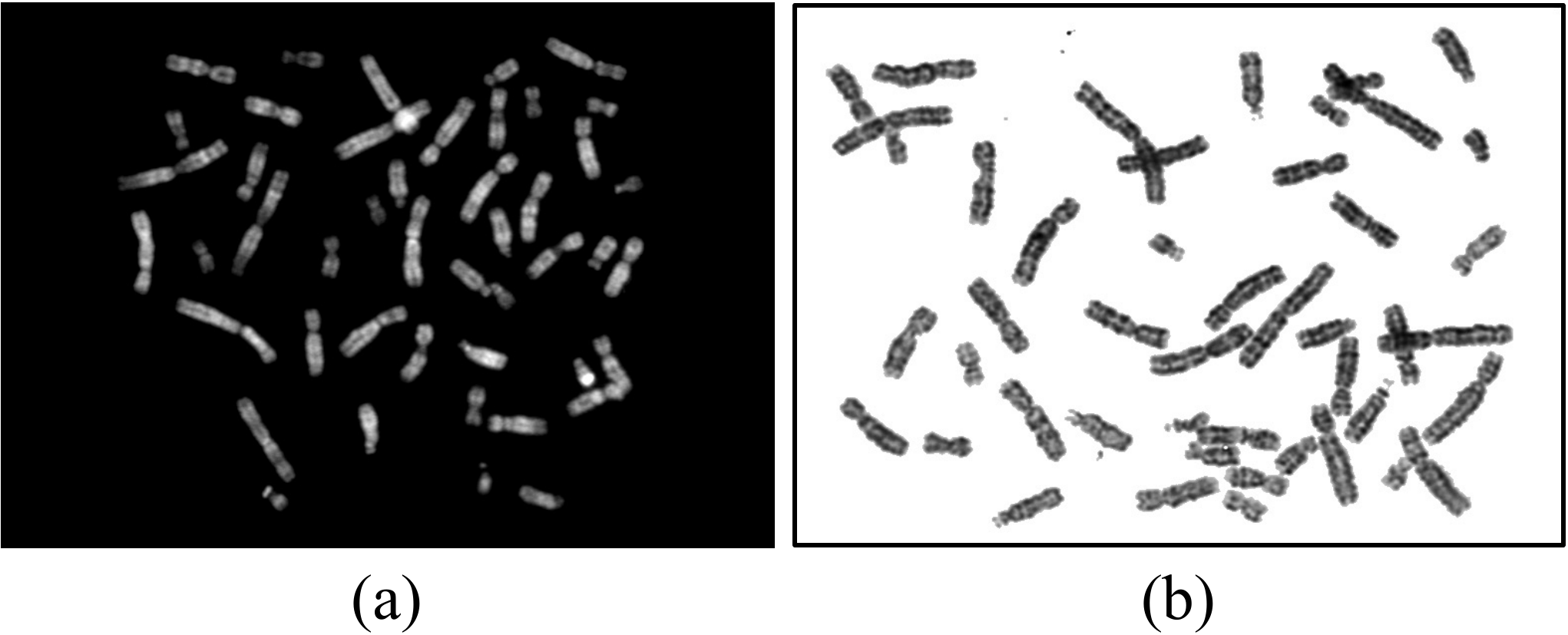}
\caption{Examples of Q-banding (a) and G-banding (b) chromosome images.}
\label{fig1}
\end{figure}

\IEEEPARstart{C}{hromosomes} are thread-like structures in the nucleus, composed of deoxyribonucleic acid molecules \cite{page2001c}. They can be numbered in pairs from 1 to 23 in each normal human cell. The first 22 pairs are known as autosomes and are arranged in descending order by length, except for chromosomes 21 and 22. The last pair of chromosomes are called the sex chromosomes and are involved in gender determination. Through the staining technique of G-banding or Q-banding \cite{maloy2013brenner}, photographic representations of condensed chromosomes become visible. Examples of chromosome images using Q-banding and G-banding are illustrated in Fig. \ref{fig1}. After acquiring stained chromosome images, Cytogeneticists place a complete set of chromosomes in a new picture as a karyotype to identify chromosomal abnormalities or genetic disorders, such as acute myeloid leukemia \cite{miyoshi1991t}, chronic lymphocytic leukemia \cite{dohner2000genomic} and Turner syndrome \cite{wolff2012erratum}. In the clinic, karyotyping is a challenging and time-consuming task that heavily relies on the expertise of cytogeneticists and is traditionally performed manually. Chromosomes can be curved in microscopic images due to their non-rigid nature. Such curvature distorts the chromosome banding pattern and further prevents cytogeneticists from analyzing chromosomes. Straightening chromosomes can address this issue by improving the readability of the banding information \cite{zhang2021chromosome} thus facilitating chromosome analysis for cytogeneticists. Furthermore, chromosome straightening has been shown to enhance the accuracy of chromosome classification \cite{sharma2017crowdsourcing, song2021novel}, ultimately leading to improved karyotyping precision. Consequently, the process of chromosome straightening is of importance in the field of chromosome analysis.


Early studies solve the chromosome straightening task using geometry \cite{jahani2012automatic,somasundaram2014straightening,roshtkhari2008novel}. Typical approaches can be divided into methods with the medial axis (MA) \cite{jahani2012automatic,somasundaram2014straightening} or bending points (BP) \cite{roshtkhari2008novel,zhang2021chromosome}. Although geometric algorithms can straighten chromosomes to some extent, interrupted or noisy banding patterns and jagged edges may exist in the straightened results. The algorithms may also cause inconsistent length due to an inaccurately extracted medial axis or incorrect bending points. Recently, advanced deep learning techniques have shown promising results in medical image analysis \cite{cui2022fully,zheng2023survival, gamper2021multiple,vonder2022deep,shui2022end} because of their strong ability in learning image representations. 
A new line of work is proposed, aiming to straighten chromosomes by establishing mapping between bent and straight chromosome using deep learning\cite{zheng2022chrsnet}. This method generates synthetic bent chromosomes by employing a non-rigid transformation strategy on real-world straight chromosomes. Generated bent and real-world straight chromosomes are then used as paired data to train a self-attention guided network. Despite reasonable straightening results obtained, the method might be restricted to straighten chromosomes with low degrees of curvature due to the difficulty in simulating highly curved chromosomes. Another line of deep learning work applies generative adversarial networks to produce straight chromosomes \cite{song2021novel,song2022robust}. In this type of method, a generator synthesizes chromosomes by the reconstruction of image representations, while a discriminator is trained to differentiate real from generated chromosomes. The advantage of utilizing generative adversarial methods is that they could generate new data closely resembling real-world chromosomes. Nonetheless, inconsistent length remains in the straightening results. The performance of the networks on straightening highly curved chromosomes also needs to be improved.

In this work, we attempt to address the issues in the aforementioned methods. We first design a preprocessing algorithm using patch rearrangement to tackle the difficulty in straightening highly curved chromosomes. To further improve the straightening performance of the preprocessing algorithm, we explore multiple conditional variational autoencoders. The autoencoders leverage curvature conditions and well learned image features to acquire straight chromosomes with satisfactory structure and banding patterns. Specifically, the preprocessing algorithm extracts consecutive rectangle patches of a chromosome along its complete medial axis. Then a curved chromosome is processed by the rearrangement of patches with their patch centers uniformly distributed in a vertical line following the same order. Such a processing step could erase low degrees of curvature in chromosomes and provide reasonable preliminary results for later straightening processes. To obtain satisfactory straightening results, we propose a novel generative model comprising a convolutional neural network, a Transformer encoder, multiple conditional variational processors, and a decoder with a Gaussian noise-based masking strategy. The convolutional neural network and transformer encoder focus on extracting local texture and global features from the chromosome image. In contrast, the conditional variational processors under the conditions of curvatures in chromosome patches reconstruct image representations that can be used by the decoder to generate an unmasked chromosome. Inspired by masked autoencoders \cite{he2022masked}, we create a reconstruction task that cannot be easily addressed by the model with a high masking ratio. Such a design helps the model to learn local and spatial features effectively. We also modify the masking with Gaussian-noise to fit the task in accordance with the image attribute of blank background in our scenario. To our knowledge, we are the first study that utilizes conditional variational autoencoders with masking for chromosome straightening. The main contributions of this work are summarized as follows:

\begin{itemize}
\item We develop a preprocessing method that can tackle the difficulty in straightening highly curved chromosomes through an accurately extracted medial axis and a strategy of patch rearrangement. The method could provide reasonable preliminary results for later straightening processes.
\item We propose a novel generative model that leverages conditional variational autoencoders with a masking strategy for chromosome straightening. The autoencoders effectively learns the mapping between banding details of chromosome patches and curvature conditions. The masking strategy enhances the feature extraction ability of the model in preserving banding patterns and structure details.
\item Extensive experiments show that our proposed framework outperforms the state-of-the-art methods on three public chromosome datasets with various stain styles. The ablation studies show the effectiveness of each component in the proposed framework.
\item Our designed framework can generate straightened chromosomes which can effectively boost the chromosome classification performance of deep learning models compared to using chromosomes without straightening, thus, allows deep learning models to provide more accurate karyotyping for cytogeneticists in clinical practice.
\end{itemize}

The remainder of the paper is organized as follows. Section II provides details of the chromosome preprocessing algorithm and the proposed masked conditional variational autoencoders. Section III presents experimental results and related analysis. Section IV describes the summary of the paper and outlines future work.

\section{Methods}

The developed framework contains two stages, namely preliminary processing and chromosome straightening. At the first stage, we cut each chromosome along its medial axis into patches. These patches are rearranged, with their patch centers uniformly distributed in a vertical line following the same center order, as the preliminary processed chromosome result. At the second stage, we propose a generative model named Masked Conditional Variational AutoEncoders (MC-VAE) to straighten the preliminary result. Next, we describe the framework in detail.

\subsection{Preliminary processing}

\begin{figure}[!t]
\centering
\includegraphics[scale=0.50]{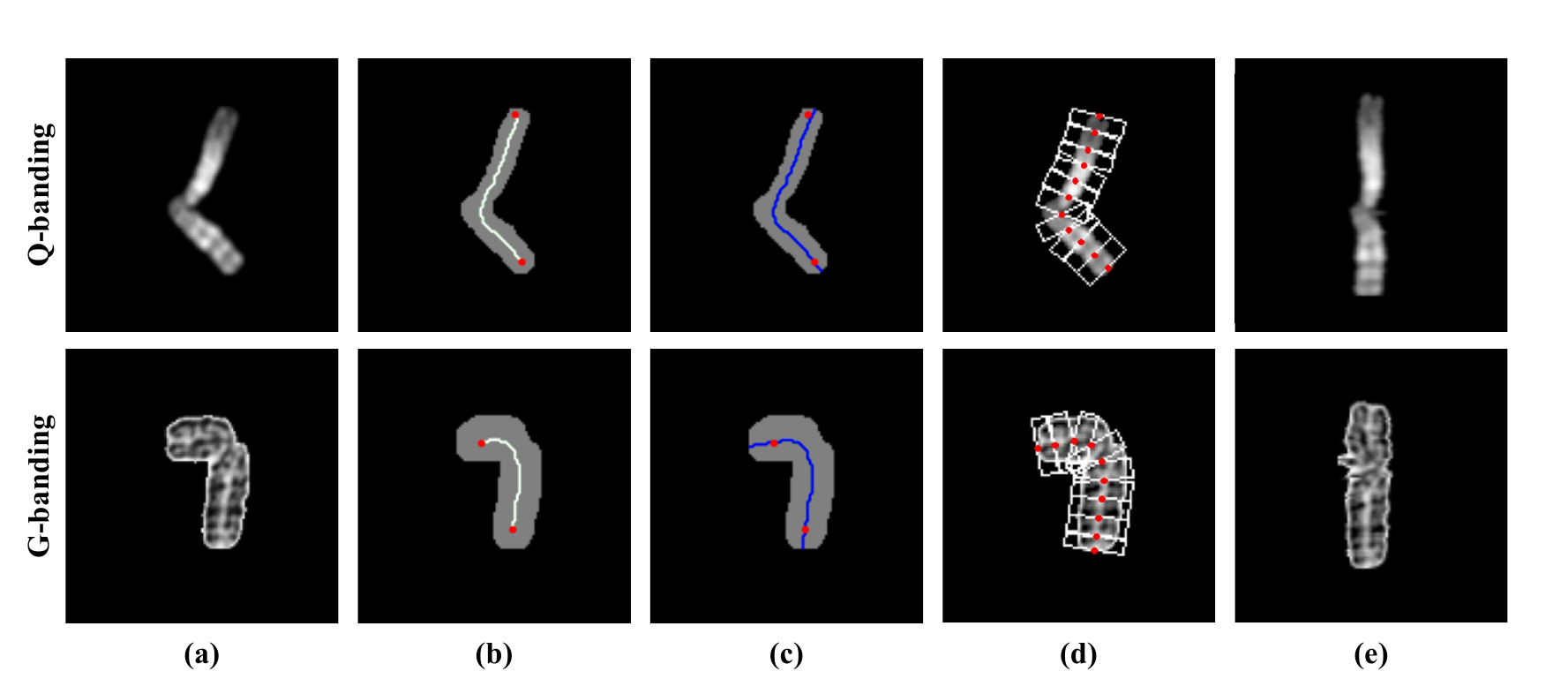}
\caption{Procedures of preliminary chromosome processing. (a) Bent chromosomes; (b) Medial axes obtained from binary chromosome images; (c) Recovered medial axes; (d) Chromosome patches extracted along the medial axis; (e) Preliminary chromosome results after rearrangement of patches.}
\label{fig2}
\end{figure}

The preliminary processing aims to erase low degrees of curvature in chromosomes to provide reasonable results for the next stage. Procedures of preliminary processing are illustrated in Fig. \ref{fig2}. Specifically, given a bent chromosome image, we first separate the chromosome from background using Otsu thresholding \cite{otsu1979threshold}. The global minimum between two main peaks in the image histogram determines the threshold. One peak is related to the background pixels and the other is corresponding to chromosome pixels. Since small variations occur in the histogram and easily affect threshold determination, we apply a median filter with the kernel size of $3\times3$ to smooth the image histogram for the acquirement of a robust threshold. Then we take the selected threshold to create a binary chromosome image. 
Holes in the chromosome area of the binary image are erased by the flood fill algorithm \cite{lu2015cfd} with four directions, followed by the utilization of the Zhang-Suen thinning algorithm \cite{zhang1984fast} with $3\times3$ search space to generate the initial medial axis of a chromosome. However, the resulting medial axis tends to be shorter than the expected length or may contain redundant branches. If the distance between the end point of the medial axis and the bottom of the chromosome exceeds 6 pixels, we recover the medial axis by iteratively calculating the local gradient around the end point and extending it until it reaches the edge. Otherwise, we first prune the medial axis with a certain ratio to remove redundant branches. Afterward, we employ the same approach as before to recover and extend the medial axis using the local gradient. 
An example of a recovered medial axis is presented in Fig. \ref{fig2}(c). The next step is to use the recovered medial axis to remove potential low degrees of curvature in chromosomes. We argue that a chromosome can be deemed as a series of rotated rectangle patches with their centers on the medial axis. Rearranging these rectangle patches can erase Low degrees of curvature. Inspired by this idea, we extract equal-sized chromosome patches and stack them without overlaps (Fig. \ref{fig2}(d)), while ensuring their centers are on the straight medial axis vertical to x-axis. The rearranged chromosome patches are used as the input at the second stage.
\subsection{Chromosome straightening}
\begin{figure*}[!t]
\centering
\includegraphics[scale=0.65]{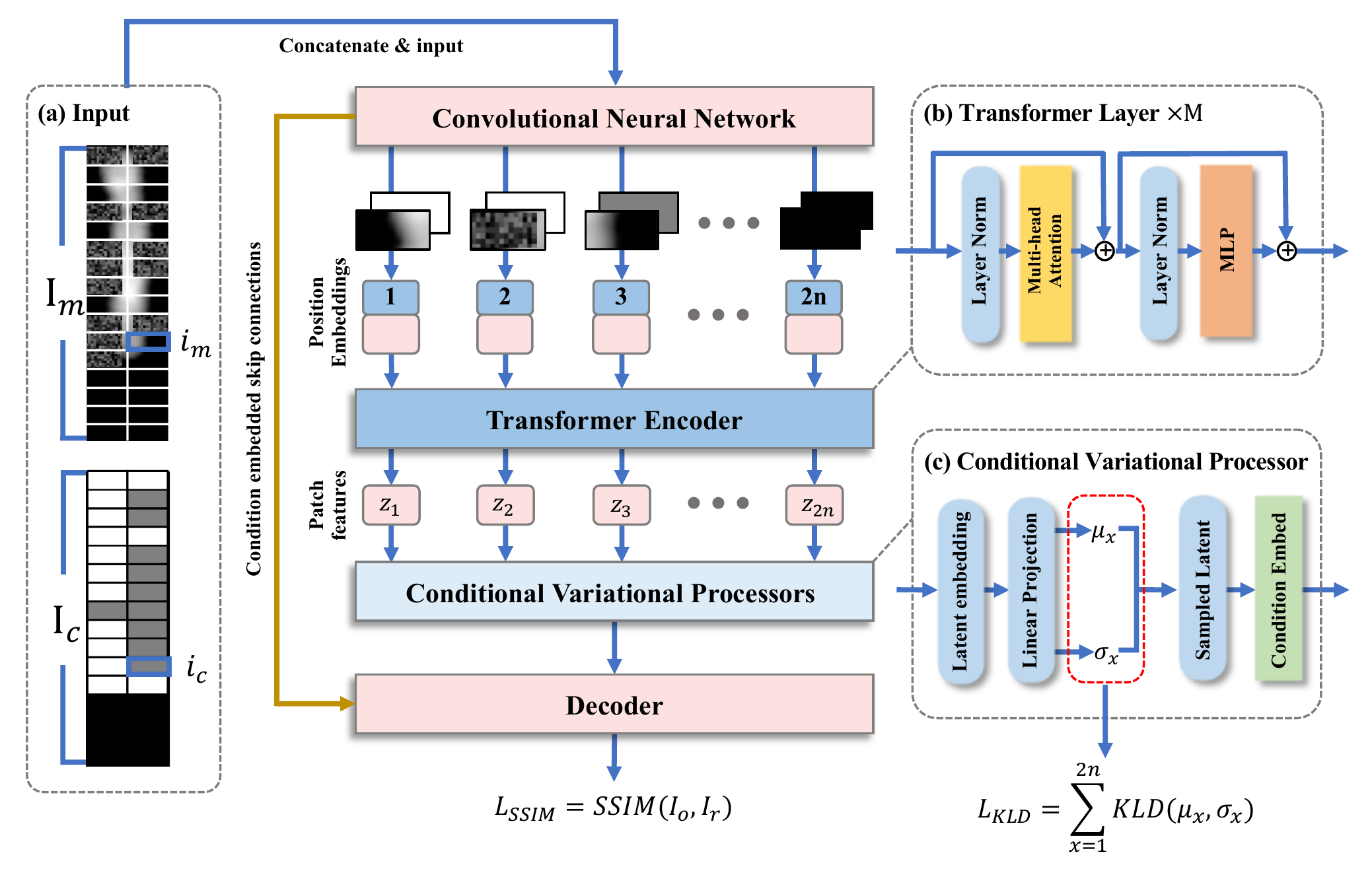}
\caption{Overview of our proposed masked conditional variational autoencoders for chromosome straightening. The model intends to reconstruct an unmasked original image $I_o$ from a masked image $I_m$ with a high masking ratio (e.g., $70\% $) according to the given condition $I_c$. $I_c$ is calculated by the Sobel operator on the binary image of $I_o$, including white, grey and black images which represent bent, straight and blank chromosome patches. A convolutional neural network is first employed to compress $I_m$ into $2n$ vectorized patches. The vectorized patches and position information are then combined as the input of the Transformer encoder where multi-head attention is applied to learn long-range relationships between chromosome patches. The output patch features $\left \{z_i | i=1,\dots,2n \right \}$ of the Transformer encoder are fed into their corresponding conditional variational processors which estimate $\mu_x$ and $\sigma_x$ by calculating the Kullback-Leibler divergence (KLD) loss between a normal distribution and a posterior distribution. At last, conditions are embedded with sampled latent vectors in the decoder to obtain the reconstructed image $I_r$. The Structural Similarity Index (SSIM) loss calculated between $I_o$ and $I_r$ is minimized to ensure the quality of $I_r$. During straightening, patches in $I_c$ are all set to grey with bent areas masked for the reconstruction of a straight chromosome.}
\label{fig3}
\end{figure*}

Our proposed masked conditional variational autoencoders (MC-VAE) aims to further straighten bent regions, to remove the image noise and to restore the missing banding patterns at the preliminary processing stage. An overview of the developed MC-VAE is illustrated in Fig. \ref{fig3}.

\subsubsection{Masking and condition}

We use the preliminarily processed chromosomes with masks as input for the training of MC-VAE. To prepare the data for training, an input image $I$ with size $H \times W$ is separated into $2n$ non-overlapping patches with size $\frac{H}{n} \times \frac{W}{2}$. Following a uniform distribution, we choose a large percentage of patches to mask and remain the rest of the patches unchanged. To keep the density profile of the medial axis during masking, the area within a distance of two pixels around the medial axis maintains the same when the masked patch has an overlap with the medial axis. This process could not only provide prior information for image reconstruction, but also is useful for later chromosome analysis, e.g., chromosome classification. An example of a masked chromosome image $I_m$ is illustrated in Fig. \ref{fig3}(a). A large ratio of masked patches can mostly remove redundant image information. Such a setting could create a task where networks use the remaining unmasked image patches to reconstruct the original image. The selection bias of masked patches can be avoided since the probability of choosing each image is equal in the uniform distribution. Unlike masked autoencoders \cite{he2022masked}, the mask in our method is not completely empty but contains Gaussian noise. The purpose of adding noise to masks is to prevent misleading from the empty background for chromosome restoration.

We employ the Sobel operator \cite{kanopoulos1988design} to produce conditional images. The operator normally utilizes two $3\times3$ kernels to estimate image gradients horizontally and vertically. In our situation, since curved edges of bent chromosomes have large horizontal gradients, we apply the Sobel operator only on every binary chromosome patch $i_b$ for approximating vertical derivatives. The condition generation process is presented as below:
\begin{equation}
g_{b}=\left[\begin{array}{ccc}
+1 & +2 & +1 \\
0 & 0 & 0 \\
-1 & -2 & -1
\end{array}\right] * i_{b}
\label{eq1}
\end{equation}
where $*$ indicates a two-dimensional convolution operation, $g_b$ is a condition patch recording the summation of gradient in $i_b$ in the horizontal direction. A large gradient means a low degree of curvature existing in $i_b$. In comparison with the value of every $g_b$, we use a threshold of $T$ to covert condition patches into white, gray and black images representing bent, straight and blank conditional labels, respectively. Consequently, a conditional image $I_c$ is presented as conditional labels with the same order and size of chromosome patches.

\subsubsection{MC-VAE}
The proposed MC-VAE is composed of a convolutional neural network, a Transformer encoder and multiple conditional variational processors (CVPs). The masked chromosome $I_m$ and its curvature condition $I_c$ are concatenated as the input of the convolutional neural network for down-sampling. The convolutional neural network comprises three ResNet blocks, of which each block is followed by a max pooling layer. The kernels of max pooling layers are varied according to the chosen size of $i_m$. After the process of the convolutional neural network, the input is converted to a sequence of D dimensional vectorized patches, where each patch has the size $\frac{H}{n} \times \frac{W}{2}$. Afterwards, we map the vectorized patches into a latent D-dimensional space using linear projections. To encode the patch spatial information, we combine patch embeddings with position embeddings $E_{pos}$ as follows:
\begin{equation}
Z_{0}=\left[x_{p}^{1} E ; x_{p}^{2} E ; \ldots ; x_{p}^{2 n} E\right]+E_{p o s}
\end{equation}
where $Z_0$ is an encoded image representation, $\left \{x_p^k |k=1,\dots,2n\right \}$ indicates $2n$ vectorized patches, and $E$ represents the patch embedding projection. The patch embedding is processed by $M$ Transformer layers, of which each layer is made of multi-head self-attention (MSA) and multi-layer perceptron (MLP) blocks. The output of a transformer layer is described as follows:
\begin{equation}
Z_{j}^{\prime}=M S A\left(L N\left(Z_{j-1}\right)\right)+Z_{j-1} \quad j=1 \ldots M
\end{equation}

\begin{equation}
Z_{j}=M L P\left(L N\left(Z_{j}^{\prime}\right)\right)+Z_{j}^{\prime}  \quad \qquad j=1 \ldots M
\end{equation}

\begin{equation}
Z_{j}=\left[z_{1, j} ; z_{2, j} ; \ldots ; z_{2n, j}\right]
\end{equation}
where $LN$ denotes the layer normalization operation, $Z_j$ is the image representation of the whole image processed by the $j_{th}$ transformer layer. It can also be presented as a set of image representations of chromosome patches.

The model has $2n$ CVPs to process the image representations from the Transformer encoder. Note that each CVP utilizes the corresponding image representation $z_x$ that already includes conditional curvature information to create a latent space. The latent space vector $h_x$ is expressed as below:
\begin{equation}
f\left(z_{x}\right)=h_{x}
\end{equation}
where $f(,)$ is a latent embedding process. In a CVP, the process using $h_x$ for reconstruction can be defined as follows:
\begin{equation}
\hat{z}_{x}=g\left(s\left(h_{x}\right)\right)
\end{equation}
where $\hat{z}_x$ is the reconstructed image representation, $s(,)$ is a sampler towards the hidden vectors, $g(,)$ is a decoding function. If $\hat{z}_x$ is directly constructed from $h_x$ by $g(,)$, minimizing the reconstruction error between $\hat{z}_x$ and $z_x$ cannot make sure that $\hat{z}_x$ and $z_x$ have the same distribution. This leads to that $g(,)$ is unable to generate the matched image representation for $z_x$ using the vector sampled from the latent space. We solve this problem by mapping the posterior distribution $p(h_x | z_x)$ of the hidden vector to a normal distribution $N(0,I)$. Then $p\left(h_x \right)$ can be expressed as:
\begin{equation}
p\left(h_{x}\right)=\sum_{x} p\left(h_{x} \mid z_{x}\right) p\left(z_{x}\right)
\end{equation}
We apply variational inference to approximate $p(h_x | z_x)$. Specifically, $\mu_x$ and $\sigma_x$ are calculated by linear projections operating on $z_x$. The sampler $s(,)$ is then used to generate the latent space vector according to the estimated mean $\mu_x$ and variance $\sigma_x$ of each vector $h_x$. During training, $\sigma_x$ equal to zero is harmful to the reconstruction process since the model could easily reconstruct the image representation without noise involved. To ensure the reconstruction ability of the model and prevent $\sigma_x$ from being zero, we employ Kullback–Leibler divergence (KLD) \cite{kingma2013auto} to make $p(h_x |z_x)$ similar to $N(0,I)$. The formula of KLD and the loss of KLD are shown below:
\begin{equation}
KLD\left[p\left(h_{x} \mid z_{x}\right) \| \mathcal{N}(0, I)\right]= \\ \frac{1}{2}\left[\left(\mu_{x}^{2}+\sigma_{x}^{2}-\log \left(\sigma_{x}^{2}\right)-1\right]\right.
\end{equation}
\begin{equation}
L_{K L D}=\sum_{x=1}^{2 n} K L \mathcal{D}\left[p\left(h_{x} \mid z_{x}\right) \| \mathcal{N}(0, I)\right]
\end{equation}
Afterwards, the condition image corresponding to $z_x$ is transformed into a vector $c_x$ concatenated with $\hat{z}_x$ as the output of a CVP. For all CVPs, the output can be written as follows:
\begin{equation}
Z_{cvps}=\left[\hat{z}_{1} c_{1} ; \hat{z}_{2} c_{2} ; \ldots ; \hat{z}_{x} c_{x}\right] \quad x=1 \ldots 2 n
\end{equation}
The decoder of MC-VAE takes $Z_{cvps}$ from CVPs as the input to reconstruct the chromosome image $I_r$. The decoder has three up-sampling blocks, of which each block contains an up-sampling layer followed by a $3\times3$ convolutional layer and a rectified linear unit. Skip-connections with embedded $I_c$ are used between decoder blocks and layers in the convolutional neural network to retain condition information at different feature map scales. We expect the reconstructed chromosome image $I_r$ is similar to the original chromosome image $I_o$. Thus, we take the Structural Similarity Index (SSIM) \cite{wang2004image} as the loss function to minimize the reconstruction error between $I_o$ and $I_r$. The SSIM value and the SSIM loss are presented below:
\begin{equation}
SSIM\left(I_{o}, I_{r}\right)=\left[\widetilde{l}\left(I_{o}, I_{r}\right)\right]^{\alpha} \cdot\left[c\left(I_{o}, I_{r}\right)\right]^{\beta} \cdot\left[r\left(I_{o}, I_{r}\right)\right]^{\gamma}
\end{equation}
\begin{equation}
L_{S S I M}=1-SSIM\left(I_{o}, I_{r}\right)
\end{equation}
where $\widetilde{l}(I_o,I_r), c(I_o,I_r), r(I_o,I_r)$ are the comparison measurements of luminance, contrast and structure between $I_o$ and $I_r$. The weight of three measurement are denoted by $\alpha$, $\beta$, and $\gamma$. The weight of $\alpha$, $\beta$, and $\gamma$ share the same value of 1. The overall loss function can be expressed as the combination of the SSIM and KLD losses:
\begin{equation}
L_{a l l}=L_{S S I M}+L_{K L D}
\end{equation}
Different from the training phase, condition patches in $I_c$ are all set to grey with bent chromosome areas masked during the straightening process so that the model produces straight chromosome patches while maintaining banding patterns. At last, produced and originally straightened chromosome patches are combined as the final result.

\subsubsection{Implementation details}
The structure of our proposed framework is implemented by PyTorch on single NVIDIA A100 GPU. The training details of MC-VAE are as follows. The MC-VAE is trained with the Adam optimizer and a batch size of 36. The learning rate is set to $10^{-3}$ with a weight decay value of $10^{-4}$. Training stops when the loss does not increase for 10 epochs, or the number of training epochs reaches 50. To generate diverse synthetic test data for training and evaluation, we randomly select 1-3 control points with a control factor ranging from 1.05 to 1.35. 

Regarding the other settings of MC-VAE, the default size ($H\times{W}$) of input images for MC-VAE is $128\times32$, the same as the patch size in the preliminary processing. It is determined by the largest chromosome in the BioImLab dataset to accommodate all chromosomes of varying sizes. When we train the model on other datasets, images of other datasets will be rescaled based on the ratio calculated using the default image size and the image size of the largest chromosome in the respective datasets. This ensures that chromosomes from other datasets can also fit in the image. Before the training of MC-VAE, each input image is separated into $32$ patches with size $8\times16$. This number of rectangle patches is determined by the optimal patch size and input image size of MC-VAE. Based on our experiments, we determine that the optimal patch size is $8\times16$ and the masking ratio is $70\%$. When generating the conditional images, the threshold of $T$ for the conversion of condition patches into different labels is equal to 18. 

For model comparisons, we include the results of our preliminary processing algorithm, MA \cite{jahani2012automatic}, BP \cite{roshtkhari2008novel} and ChrSNet \cite{zheng2022chrsnet}. The implementation details of these algorithms are as follows. The training of ChrSNet follows the settings in \cite{zheng2022chrsnet}. An initial learning rate of $5\times10^{-4}$ is set and gradually decreased to $10^{-5}$. The optimizer used is Adam, and the batch size is 24. The weights used to determine the rotation score in \cite{roshtkhari2008novel} are all set to 0.5 when applying the BP method. As for the MA method, we employ the Zhang-Suen thinning algorithm \cite{zhang1984fast} with a $3\times3$ search space to generate medial axes for chromosomes. The medial axes are extended by 30 pixels at both ends using the slope calculated by the last 5 pixels. Regarding the preliminary processing algorithm, the rotated angle for patches in Fig. \ref{fig2}(d) is calculated based on the gradient between the center red points of the patches. The patch size and prune ratio are $8\times16$ and 0.1, respectively. 

\section{Experiments and results}
\subsection{Datasets}
The proposed framework is evaluated on public datasets. Since curvatures are normally observed on chromosomes with long arms, we focus on the straightening of chromosomes 1-12 in these three datasets. We divide real-world chromosomes in each dataset into bent and straight ones according to the MA score. Straight chromosomes are used to produce five synthetic bent chromosomes by employing a non-rigid data generation strategy proposed in \cite{zheng2022chrsnet} to expand the data for model development. Then real-world and synthetic bent chromosomes are used to train the framework. In our study, a five-fold cross validation is deployed for experiments. Random separation of real-world data and its corresponding synthetic data into five subsets might lead to the situation that both data occur in training and testing subsets at the same time. This situation can cause overestimated performance of deep learning models for chromosome straightening. Therefore, during the data splitting process, we include real-world data and its corresponding synthetic data in the same subsets to prevent information leakage.
Besides, performing evaluation on synthetic data in addition to real-world data can further show the model effectiveness \cite{liu2020no,lin2021learning}. So, we generate diverse synthetic data and evaluate our model on it.


Two datasets with different stains are used for model comparison and ablation study. The first dataset, BioImLab \cite{poletti2008automatic}, consists of Q-band stained chromosomes. We include 694 real-world bent and 500 straight chromosomes for experiments. The other  dataset Pki \cite{ritter2008automatic} is made of G-band stained chromosomes. We conduct experiments on 3,554 real-world bent and 3,000 straight chromosomes.

Another dataset of ChromosomeNet \cite{lin2022chromosomenet} including G-band stained  images in metaphase is served as an independent unseen dataset. We select 1,633 real world bent and 1,500 straight chromosomes from this dataset for experiments. 


\subsection{Evaluation metrics}

We evaluate the performance of the framework in respect of structure features and banding patterns. The metric of length is defined as follows:
\begin{equation}
L\ score = \left(1-\frac{\left|l-l^{\prime}\right|}{l^{\prime}}\right) * 100
\end{equation}
where $l$ and $l'$ are the predicted length and the target length calculated by counting the number of pixels in the chromosome skeleton. Straightness of chromosome is evaluated in both global and local ways. The global straightness score regarding the medial axis (MA) is calculated by the slope variations between the overall slope and local slopes. The local slopes are calculated by n points uniformly sampled on the medial axis. The MA score can be written as follows:
\begin{equation}
MA\ score=\left(1-\frac{1}{n} \sum_{i=1}^{n}\left[\frac{y_{i}-y_{i-1}}{x_{i}-x_{i-1}}-\frac{y_{b}-y_{t}}{x_{b}-x_{t}}\right]\right) * 100
\end{equation}
where the coordinate of the $i_{th}$ point $(x_i,y_i)$ is sampled between the two points $(x_b,y_b)$ and $(x_t,y_t)$ at the ends of a chromosome. The number of sample points $n$ is set to 6. The Sobel operator is applied to evaluate the straightness of chromosomes locally. The straightness score calculated by the Sobel operator can be defined as below:
\begin{equation}
Sobel\ score=\lambda \sum_{x=1}^{2 n} K * i_{b}
\end{equation}
where the overall Sobel score is the summation of gradients of $2n$ chromosome patches. $K$ is the kernel for the calculation of the horizontal gradients and $\lambda$ is a scale factor of $10^{-2}$.
Banding patterns affect not only the classification of chromosomes, but also the identification of chromosomal abnormalities. Therefore, it is of importance to evaluate the banding patterns after chromosome straightening. Due to morphological deformation between curved and straightened chromosomes, Structural Similarity Index Measure (SSIM) \cite{wang2004image} might be inappropriate for the evaluation of pattern preservation. Compared to SSIM, Learned Perceptual Image Patch Similarity (LPIPS) \cite{zhang2018unreasonable} is a more neutral metric to calculate the perceptual distance between input and output images. Hence, we adopt LPIPS to estimate the quality of banding features after the straightening process. Besides, density profile records pixel values along the medial axis of the chromosome and associates with the type of chromosomes \cite{moradi2006new}. Therefore, density profile is also applied to assess the consistency of banding patterns before and after the restoration. The score of the density profile (DP) is presented as below:
\begin{equation}
DP\ score =\sum_{i=1}^{N}\left(y_{i}-y_{i}^{\prime}\right)^{2}
\end{equation}
where $y_i$ and $y_i'$ are the $i_{th}$ pixels on the medial axes of a bent chromosome and a straightened chromosome, respectively. And $N$ represents the number of pixels on the medial axis of $y_i$.

To further verify the effectiveness of our framework in chromosome straightening, we also assess the impact of implementing our framework in the task of chromosome classification by five-fold cross-validation. Since chromosome images normally have low resolution, increasing the complexity of a deep learning model might result in decreased accuracy \cite{song2021novel,song2022robust}. Hence, we apply VGG-16 \cite{simonyan2014very}, ResNet-18 \cite{he2016deep}  and AlexNet \cite{krizhevsky2017imagenet} for experiments. Accuracy, precision, recall, and $F_1$ values are used as metrics, defined as follows:
\begin{equation}
accuracy=\frac{1}{C} \sum_{j=1}^{C} \frac{TP_{j} + TN_j}{TP_j + TN_j+FP_j +FN_j}
\end{equation}
\begin{equation}
recall=\frac{1}{C} \sum_{j=1}^{C} \frac{T P_{j}}{T P_{j}+F N_{j}}
\end{equation}
\begin{equation}
precision=\frac{1}{C} \sum_{j=1}^{C} \frac{T P_{j}}{T P_{j}+F P_{j}}
\end{equation}
\begin{equation}
F_1=\frac{2 \times precision \times recall }{ precision + recall }
\end{equation}
where $C$ is the number of chromosome types. $TP$, $FP$, $TN$ and $FN$ represent true positive, false positive, true negative and false negative cases separately.

\subsection{Model comparisons}

\begin{table*}[!h]
\centering
\caption{Performance comparisons between methods for chromosome straightening on the BioImLab dataset. BP indicates the geometric method with bending points, while MA represents the geometric approach using medial axes. PPA is our proposed preliminary processing algorithm that is a modified MA method. ChrSNet, and ours including PPA and MC-VAE use deep learning techniques. The best result in each column is shown in bold.}
\begin{tabular}{llllll}
\hline
\textbf{Method} & \textit{\textbf{L\ score}}$\uparrow$ & \textit{\textbf{MA score}} $\uparrow$ & \textit{\textbf{Sobel score}} $\downarrow$  & \textit{\textbf{LPIPS}} $\uparrow$ & \textit{\textbf{DP score}} $\downarrow$\\ \hline
Synthetic data                       &             &             &              &             &             \\
\quad BP \cite{roshtkhari2008novel}       & $83.51\pm8.94$ & $77.64\pm7.44$ & $194.54\pm1.24$ & $86.24\pm5.42$ & $81.65\pm6.40$ \\
\quad MA \cite{jahani2012automatic}       & $94.86\pm2.91$ & $94.65\pm2.96$ & $122.67\pm3.29$ & $91.28\pm2.04$ & $63.76\pm4.21$ \\

\quad PPA      & $94.88\pm0.79$ & $95.81\pm0.81$ & $111.25\pm3.14$ & $92.02\pm1.17$ & $40.93\pm5.47$ \\

\quad ChrSNet \cite{zheng2022chrsnet}     & $95.14\pm5.58$ & $88.94\pm3.84$ & $139.77\pm6.28$ & $89.25\pm3.71$ & $68.82\pm6.09$ \\

\quad ChrSNet \cite{zheng2022chrsnet}+PPA    &$ 95.18\pm6.69 $&$ 91.52\pm4.78 $ & $118.40\pm7.59$ &$ 90.64\pm2.83 $& $53.32\pm7.38$ \\
\quad MC-VAE    &$ 95.15\pm1.01 $&$ 95.13\pm5.14 $ & $93.29\pm1.18$ &$ 91.93\pm1.09 $& $42.13\pm4.13$ \\

\quad Ours                                 & $\textbf{96.24}\pm\textbf{0.42}$ & $\textbf{97.45}\pm\textbf{0.14}$ & $\textbf{84.52}\pm\textbf{4.50}$  & $\textbf{93.22}\pm\textbf{0.50}$ & $\textbf{27.92}\pm\textbf{2.08}$ \\
Real-world data                      &             &             &              &             &             \\
\quad BP \cite{roshtkhari2008novel}       & $85.98\pm7.32$ & $82.58\pm9.98$ & $176.00\pm9.02$ & $86.92\pm5.93$ & $74.43\pm3.63$ \\
\quad MA \cite{jahani2012automatic}       & $90.45\pm2.35$ & $92.14\pm2.91$ & $143.91\pm4.75$ & $91.65\pm2.82$ & $51.84\pm2.37$ \\

\quad PPA      & $90.55\pm0.86$ & $96.53\pm0.97$ & $127.61\pm5.85$ & $91.67\pm1.49$ & $35.59\pm4.93$ \\

\quad ChrSNet \cite{zheng2022chrsnet}  & $93.87\pm3.72$ & $88.25\pm5.13$ & $165.93\pm7.58$ & $84.95\pm4.24$ & $62.60\pm5.29$ \\

\quad ChrSNet \cite{zheng2022chrsnet}+PPA    &$ 93.88\pm5.21 $&$ 93.29\pm4.96 $ & $129.89\pm6.76$ &$ 88.92\pm3.91 $& $51.67\pm5.18$ \\
\quad MC-VAE    &$ 93.91\pm2.36 $&$ 94.92\pm0.87 $ & $99.19\pm1.29$ &$ 92.33\pm0.85 $& $35.71\pm1.83$ \\

\quad Ours                      & $\textbf{94.63}\pm\textbf{1.95}$ & $\textbf{97.24}\pm\textbf{0.12}$ & $\textbf{87.97}\pm\textbf{0.75}$  & $\textbf{93.41}\pm\textbf{0.43}$ & $\textbf{27.41}\pm\textbf{0.86}$ \\ \hline
\end{tabular}
\label{t1}
\end{table*}

\begin{table*}[!h]
\centering
\caption{Performance comparisons between methods for chromosome straightening on the Pki dataset. The best result in each column is shown in bold.}
\label{t2}
\begin{tabular}{llllll}
\hline
\textbf{Method} & \textit{\textbf{L\ score}}$\uparrow$ & \textit{\textbf{MA score}} $\uparrow$ & \textit{\textbf{Sobel score}} $\downarrow$  & \textit{\textbf{LPIPS}} $\uparrow$ & \textit{\textbf{DP score}} $\downarrow$\\ \hline
Synthetic data                       &             &             &              &             &             \\
\quad BP \cite{roshtkhari2008novel}       &$ 84.10\pm5.26$ &$ 66.35\pm13.47$ & $312.39\pm17.24$ &$ 84.33\pm7.19$ & $186.22\pm8.07$ \\
\quad MA \cite{jahani2012automatic}       &$ 86.49\pm2.14$ &$ 88.13\pm3.17$  & $135.78\pm8.97$  &$ 89.96\pm2.27$ & $120.75\pm4.12$ \\
\quad PPA      &$ 86.59\pm3.45 $&$ 88.24\pm2.98 $& $127.81\pm13.35$ &$ 90.01\pm3.23 $&$ 98.69\pm6.27 $  \\
\quad ChrSNet \cite{zheng2022chrsnet}     &$ 91.51\pm4.95$ &$ 91.93\pm7.23$  &$ 93.34\pm6.29$   &$ 92.37\pm3.04$ &$ 95.27\pm9.22$  \\

\quad ChrSNet \cite{zheng2022chrsnet}+PPA     &$91.75\pm4.63$&$91.98\pm6.37$ &$93.29\pm5.13$ &$92.51\pm3.16$	 &$93.17 \pm 8.33$   \\
\quad MC-VAE     &$ 90.58\pm1.86$ & $90.34\pm1.96 $& $92.53\pm5.03$ &$ 91.29\pm1.44 $&$ 90.16\pm5.43 $  \\

\quad Ours                                &$ \textbf{92.28}\pm\textbf{1.20} $ &$ \textbf{92.03}\pm\textbf{1.21}$  &$ \textbf{88.32}\pm\textbf{8.92}$   &$ \textbf{92.75}\pm\textbf{0.91}$ &$ \textbf{89.76}\pm\textbf{1.97}$  \\

Real-world data                      &             &             &              &             &             \\
\quad BP \cite{roshtkhari2008novel}       &$ 80.41\pm9.58$ &$ 62.05\pm12.86$ & $291.49\pm8.14$  &$ 81.43\pm4.58$ & $179.14\pm7.05$ \\
\quad MA \cite{jahani2012automatic}       &$ 82.98\pm3.45$ &$ 84.29\pm4.91$  & $144.56\pm9.46$  &$ 88.34\pm3.37$ & $138.69\pm6.21$ \\

\quad PPA      &$ 83.15\pm3.41 $&$ 84.67\pm3.64 $& $148.85\pm8.73$ &$ 88.45\pm3.46 $& $99.85\pm3.36 $ \\

\quad ChrSNet \cite{zheng2022chrsnet}     &$ 85.51\pm5.09$ &$ 78.23\pm14.06$ & $223.79\pm7.27$  &$ 88.36\pm3.05$ & $119.58\pm8.36$ \\

\quad ChrSNet \cite{zheng2022chrsnet}+PPA     &$85.58\pm6.28$&$82.41\pm9.03$ &$134.64\pm8.85$ &$88.74\pm3.73$	 &$102.48\pm11.94$ \\
\quad MC-VAE    &$ 83.16\pm1.16 $&$ 85.03\pm3.82 $& $143.32\pm13.32$ &$ 89.20\pm5.69 $&$ 97.56\pm7.11 $  \\

\quad Ours                                 &$ \textbf{86.03}\pm\textbf{2.10}$ &$ \textbf{90.53}\pm\textbf{1.10}$  &$ \textbf{94.18}\pm\textbf{7.81}$   &$ \textbf{90.37}\pm\textbf{1.13}$ &$ \textbf{96.54}\pm\textbf{1.72}$  \\ \hline
\end{tabular}
\end{table*}

We compare the performance of geometric and deep learning methods on synthetic and real-world data acquired from BioImLab and Pki dataset for chromosome straightening. The results of comparisons are given in Table \ref{t1}-\ref{t2}. After model evaluation, our method achieved favorable results not only on synthetic chromosomes, but also on real-world data. This can show that our method is effective in chromosome straightening. Specifically, we can find that our proposed framework outperforms the other methods by a large margin on two datasets from the tables. In contrast to the $L$ scores achieved by \cite{jahani2012automatic}, our framework improves the results by $1.38$, $5.79$ on the synthetic data, and $4.18$, $3.05$ on the real-world data. This shows the effectiveness of our framework in maintaining chromosome length. When considering the straightness of the medial axis, the framework yields a score of $97.45$, $97.24$, $92.03$, $90.53$ on the BioImLab synthetic, BioImLab real-world, Pki synthetic, Pki real-world data, respectively. The high score means chromosomes is well straightened globally. The Sobel score is calculated to evaluate the straightness of chromosomes at the edge. Comparing with \cite{roshtkhari2008novel}, the Sobel score of our method halved on the BioImLab dataset and decreased more than a factor of three on the Pki dataset. The improvement in the Sobel score might suggest that our method is able to keep smooth edges for straightened chromosomes. In terms of LPIPS, a deep learning method ChrSNet without our preliminary processing algorithm has $84.95 \pm 4.24$, and $88.36\pm 3.05$ on the real-world data of the two datasets, whereas our framework reaches $93.41\pm 0.43$, and $90.37\pm 1.13$. The larger mean values and smaller standard deviations of our method in comparison with ChrSNet (without PPA) might indicate the superiority of our method in preserving banding patterns. To assess the pattern consistency at the medial axis, we also compare the DP scores between methods. We can observe that our method achieves the lowest DP scores ($27.92$, $27.41$, $89.76$, $96.54$) among methods, which shows that the density profile can be accurately restored. To determine the benefits that MC-VAE brings to chromosome straightening on the preliminary processed image, we also evaluate the performance of the preliminary processing algorithm. By comparing the results in the "PPA" and "Ours" rows, we can find that MC-VAE is able to bring further improvement to the preliminary processed images in terms of structure consistency and banding features. This further shows the capability of MC-VAE for chromosome straightening.
Furthermore, we explore if MC-VAE with PPA can have better performance than that of ChrSNet with PPA. From Table \ref{t1}-\ref{t2}, we can find that although the performance of ChrSNet is improved by using PPA, its performance is still inferior to the combination of PPA and MC-VAE (Ours). This shows the superiority of MC-VAE in enhancing the straightening performance based on the preliminary results of PPA.
Besides, we conduct the experiments with the original chromosomes as the input of MC-VAE. We find that even without using preliminary processing on the chromosomes, MC-VAE still produces favorable results on chromosome straightening. It can achieve better-preserved structure and banding features than the MA and BP methods. But the performance is worse than that of the framework (Ours) using preprocessed chromosomes. This provides evidence that the utilization of preliminary processing is beneficial to improve the performance of MC-VAE.

\begin{figure*}[!tb]
\centering
\includegraphics[scale=0.70]{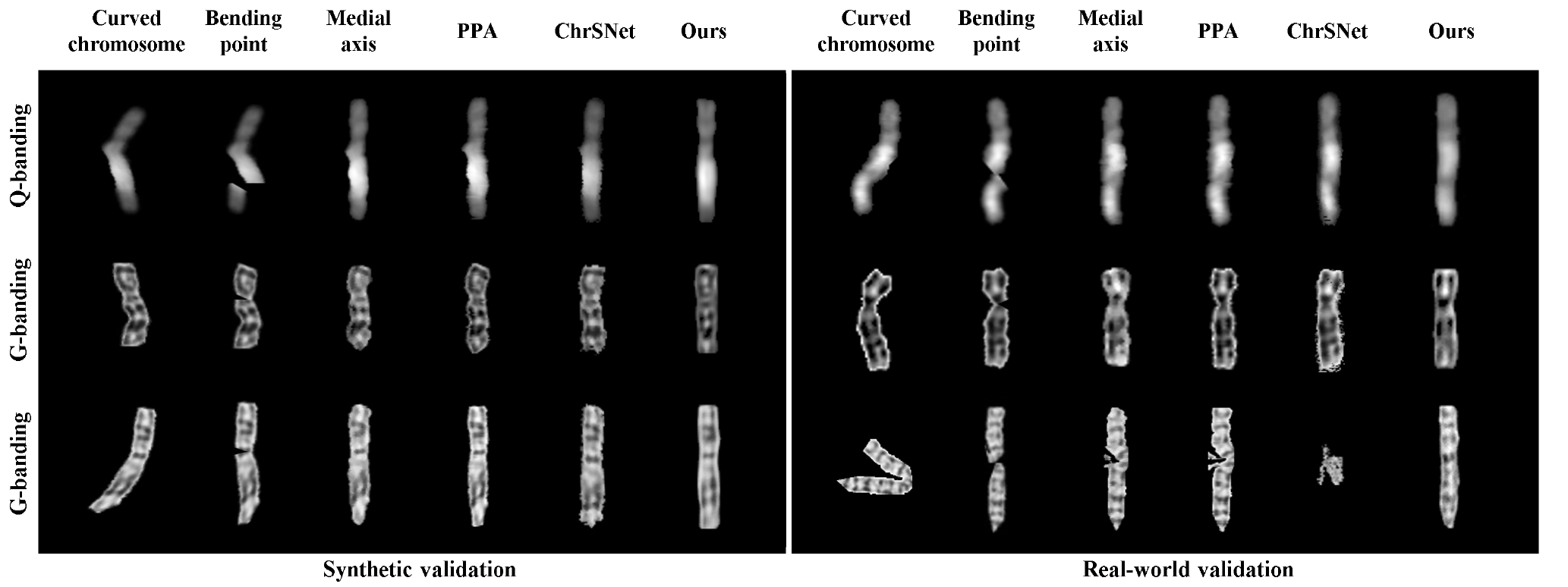} %
\caption{\label{fig4} Examples of chromosomes straightened by different methods. Compared to other methods, our proposed approach could effectively erase the curvatures of different chromosomes meanwhile maintaining structure details and banding patterns on synthetic and real-world validation.}
\end{figure*}

\begin{table*}[!tb]
\centering
\caption{Ablation study of the proposed method on the Pki dataset. PPA, BP, MA, CNN, CVPs, Trans, BRM, GNLRM, and GNRM indicate the preliminary processing algorithm, bending point method, medial axis method, convolutional neural network, conditional variational processors, Transformer encoder, blank random masking, Gaussian noise-based left or right masking, and Gaussian noise-based random masking in all patches, respectively.}
\label{t3}
\begin{tabular}{llllll}
\hline
\textbf{Components} & \textit{\textbf{L\ score}}$\uparrow$ & \textit{\textbf{MA score}} $\uparrow$ & \textit{\textbf{Sobel score}} $\downarrow$  & \textit{\textbf{LPIPS}} $\uparrow$ & \textit{\textbf{DP score}} $\downarrow$\\ \hline
Synthetic data     &             &             &               &             &               \\
\quad PPA      &$ 86.59\pm3.45 $&$ 88.24\pm2.98 $& $127.81\pm13.35$ &$ 90.01\pm3.23 $&$ 98.69\pm6.27 $    \\
\quad PPA+CNN+Trans+GNRM     &$ 86.85\pm1.02 $&$ 87.92\pm2.80 $&$ 134.2\pm26.55$  &$ 91.97\pm2.12 $&$ 96.38\pm6.85 $  \\
\quad PPA+CNN+CVPs+GNRM      &$ 87.32\pm3.30 $&$ 88.12\pm3.94 $&$ 97.69\pm12.24$  &$ 90.21\pm4.40 $& $100.13\pm13.25$ \\
\quad PPA+CNN+Trans+CVPs+BRM       &$ 89.10\pm2.89 $&$ 91.06\pm2.18 $&$ 93.76\pm0.64 $  &$ 91.65\pm3.26 $&$ 95.15\pm5.25 $  \\

\quad CNN+Trans+CVPs+GNRM    &$ 90.58\pm1.86$ & $90.34\pm1.96 $& $92.53\pm5.03$ &$ 91.29\pm1.44 $&$ 90.16\pm5.43 $  \\

\quad PPA+CNN+Trans+CVPs+GNLRM  &  $87.84\pm4.76$	&         
$89.45\pm1.68$	&         $98.1\pm9.63$ &        $91.89\pm2.95$	&         $90.63\pm6.36$	\\

\quad BP\cite{roshtkhari2008novel}+CNN+Trans+CVPs+GNRM  &$ 88.27\pm3.51$ &$ 80.67\pm2.35$  &$ 186.53\pm19.85$&$ 88.97\pm6.29$ &$ 137.96\pm8.57$  \\
\quad MA\cite{jahani2012automatic}+CNN+Trans+CVPs+GNRM  &$ 90.93\pm2.29$ &$ 88.49\pm2.70$  &$ 103.42\pm11.53$   &$ 90.87\pm3.98$ &$ 94.89\pm8.36$  \\

\quad PPA+CNN+Trans+CVPs+GNRM &$ \textbf{92.28}\pm\textbf{1.20} $&$ \textbf{92.03}\pm\textbf{1.21} $&$ \textbf{88.32}\pm\textbf{8.92} $  &$ \textbf{92.75}\pm\textbf{0.91} $&$ \textbf{89.76}\pm\textbf{1.97} $  \\

Real-world data    &             &             &               &             &               \\
\quad PPA      &$ 83.15\pm3.41 $&$ 84.67\pm3.64 $& $148.85\pm8.73$ &$ 88.45\pm3.46 $& $99.85\pm3.36 $ \\
\quad PPA+CNN+Trans+GNRM     &$ 83.16\pm1.16 $&$ 85.03\pm3.82 $& $143.32\pm28.32$ &$ 89.20\pm5.69 $&$ 97.56\pm7.11 $  \\
\quad PPA+CNN+CVPs+GNRM      &$ 83.86\pm3.85 $&$ 85.68\pm3.11 $& $102.09\pm22.19$ &$ 88.60\pm4.85 $& $101.34\pm2.35 $ \\
\quad PPA+CNN+Trans+CVPs+BRM       &$ 84.25\pm0.69 $&$ 90.34\pm2.70 $&$ 98.74\pm1.99 $  &$ 89.04\pm3.46 $& $105.79\pm4.99 $ \\
\quad CNN+Trans+CVPs+GNRM &$ 84.06\pm4.13 $&$ 86.08\pm1.65 $& $102.41\pm7.79$ &$ 89.25\pm1.46 $&$ 100.77\pm3.96 $  \\

\quad PPA+CNN+Trans+CVPs+GNLRM  &  $83.91\pm3.75$	&      $89.35\pm2.88$	&    $108.62\pm10.29$&       $89.24\pm2.14$	&         $97.08\pm4.23$	        \\

\quad BP\cite{roshtkhari2008novel}+CNN+Trans+CVPs+GNRM  &$ 82.21\pm2.40$ &$ 72.48\pm13.59$  &$ 170.31\pm12.46$   &$ 86.53\pm19.85$ &$ 120.96\pm8.57$  \\
\quad MA\cite{jahani2012automatic}+CNN+Trans+CVPs+GNRM  &$ 84.95\pm4.04$ &$ 86.89\pm3.92$  &$ 128.63\pm9.26$   &$ 89.17\pm2.89$ &$ 113.25\pm7.91$  \\

\quad PPA+CNN+Trans+CVPs+GNRM&$ \textbf{86.03}\pm\textbf{2.10} $&$ \textbf{90.53}\pm\textbf{1.10} $&$ \textbf{94.18}\pm\textbf{7.81} $  &$ \textbf{90.37}\pm\textbf{1.13} $&$ \textbf{96.54}\pm\textbf{1.72} $\\ 

\hline
\end{tabular}
\end{table*}

To visualize the straightened results, we present examples of chromosomes processed by different methods in Fig. \ref{fig4}. The geometric method using bending points (BP) \cite{roshtkhari2008novel} could erase the curvature of chromosomes, but banding details close to the centromere are easily lost. The medial axis based method \cite{jahani2012automatic} could solve the problem of the BP method to some extent by vertically sampling along the medial axis of the chromosome. A side effect of using the BP, MA or PPA methods is that discontinuous banding patterns and jagged edges may be introduced, which can prevent the later analysis of chromosomes. Despite the good performance on straightening chromosomes in the first two rows, ChrSNet fails to straighten the chromosome with low degrees of curvature in the third row on real-world validation. This might be attributed to the difficulty of ChrSNet in simulating highly curved chromosomes for model training. Among five chromosome straightening methods, our designed framework produces straightened chromosomes that are closest to the curved ones in terms of banding features. This may indicate the effectiveness of the proposed MC-VAE in erasing curvatures of chromosomes meanwhile maintaining structure details and banding patterns for chromosome straightening.

\subsection{Ablation studies}

\begin{figure*}[!tb]
\centering
\includegraphics[scale=0.58]{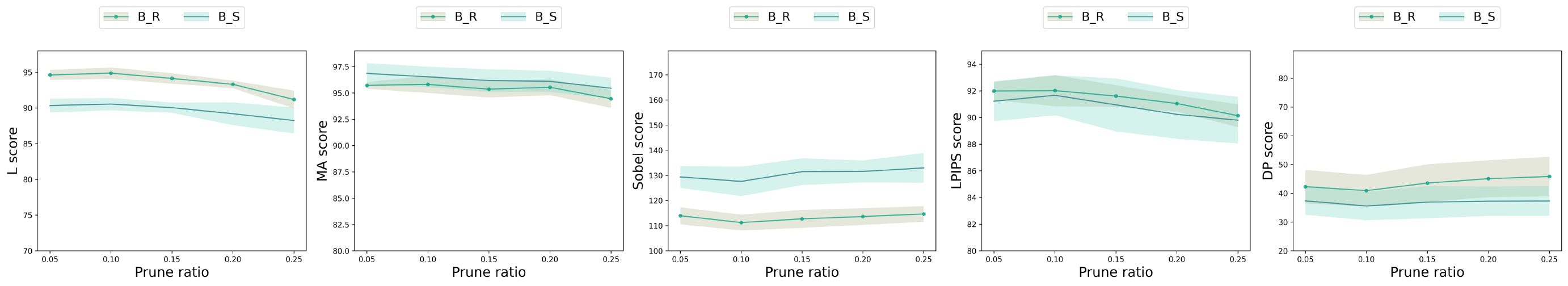}
\caption{\label{fig5} Effect of prune ratios on the performance of our preliminary processing algorithm. The algorithm with the prune ratio of $0.10$ results in good performance on the real-world (R) and synthetic (S) chromosomes from BioImLab dataset.}
\end{figure*}

\begin{figure*}[!tb]
\centering
\includegraphics[scale=0.58]{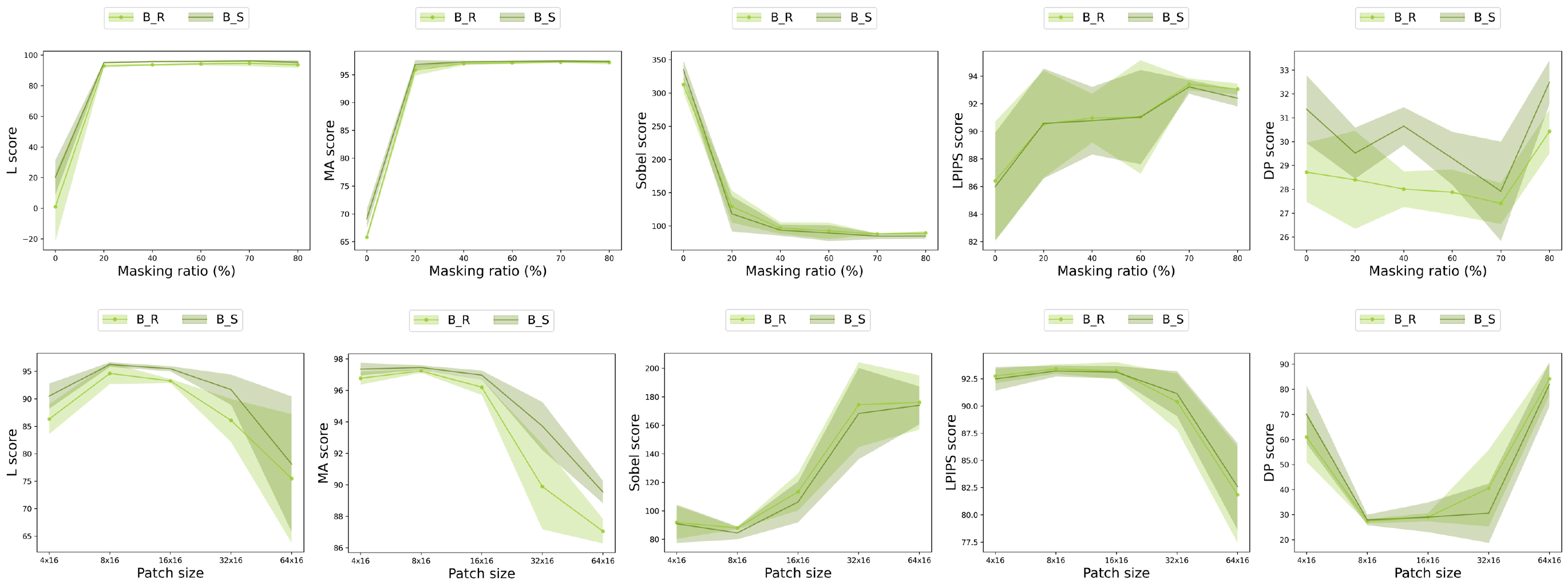}
\caption{\label{fig6} Effect of masking ratios (up) and patch sizes (down) on the performance of MC-VAE. A high masking ratio ($70\%$) and a small patch with size ($8\times16$) results in good performance on the real-world (R) and synthetic (S) chromosomes from BioImLab dataset.}
\end{figure*}

We investigate the effect of key components in the proposed MC-VAE on the synthetic and real-world data from the largest chromosome dataset of Pki, as shown in Table \ref{t3}. The baseline results are generated by using the preliminary preprocessing algorithm (PPA) only. The effectiveness of each component is evaluated by gradually adding or replacing modules. Comparing with the result of PPA, the combination of PPA, CNN and the Transformer encoder with Gaussian noise-based random masking (GNRM) achieves better results on the real-world data in terms of the L, MA, Sobel, LPIPS and DP scores with a value of $83.16$, $85.03$, $143.32$, $89.20$, $97.56$, respectively. This might suggest the effectiveness of hybrid encoder for feature extraction and image reconstruction. However, the use of PPA, CNN, CVPs and GNRM acquires better L, MA and Sobel scores ($83.86$, $85.68$, $102.09$) than those ($83.16$, $85.03$, $143.32$) of the utilization of PPA, CNN, GNRM and the Transformer encoder, which shows that the former is more effective to maintain structure details in chromosome straightening than the latter one. The fusion of PPA, CNN, CVPs and the Transformer encoder with blank random masking results in an improvement regarding the L, MA, and Sobel score but a degradation in terms of LPIPS, compared to the aggregation of PPA, CNN, GNRM and the Transformer encoder. These results might indicate that the structure information of chromosomes can be better restored by the fused model, but the preservation of banding patterns can still be optimized. This can be mainly attributed to the use of blank masks which confuses the model for feature extraction of banding patterns. The model cannot differentiate between blank masks and blank background. Thus, we introduce Gaussian noise for masking in the previous experiments. The results in the fourth (PPA+CNN+Trans+CVPs+BRM) and last rows (PPA+CNN+Trans+CVPs+GNRM) show that GNRM can enhance the ability of the model for chromosome straightening. 
Based on Gaussian noise-based masking, we further explore the effect of randomly masking left or right patches with Gaussian noise (GNLRM) on each row of chromosome images. By comparing the results in the rows (PPA+CNN+Trans+CVPs+GNLRM and PPA+CNN+Trans+CVPs+GNRM), we can find that the model performs better with GNRM used in all patches. The masking ratio setting is $70\%$ when GNRM is applied in all patches, whereas the maximum masking ratio achievable with GNLRM is $50\%$. There is still redundant information that can be removed to enhance the model's feature learning. This finding aligns with our experiments on the effect of masking ratios ($0\%$ to $80\%$) on model performance in the next subsection. When the masking ratio is below the optimal value, preserving more redundant information leads to worse model performance. To further show PPA brings more benefits to MC-VAE (consisting of CNN+Trans+CVPs+GNRM) compared to other geometric methods, we conduct experiments to train MC-VAE using the results obtained from PPA and the others. By comparing the results of MA, BP, PPA combined with MC-VAE, we observe that the combination of PPA and MC-VAE performs the best. This shows that utilizing PPA is beneficial in enhancing the performance of MC-VAE. The final approach taking the key components of PPA, CNN, the Transformer encoder, CVPs and GNRM achieves the best performance regardless of metrics for structure or banding patterns. The results in Table \ref{t3} show the effectiveness of each component in our proposed approach.

\subsection{Effect of parameters}

\begin{table}[!t]
\centering
\caption{Performance for chromosome classification using original and straightened chromosomes produced by multiple methods on the BioImLab dataset. }
\label{t4}
\begin{tabular}{lllll}
\hline
\textbf{Model}                       & \textbf{Accuracy} & \textbf{Precion} & \textbf{Recall} & \textbf{$F_1$ Score} \\ \hline
VGG-16                               &                   &                    &                 &                   \\
\quad Origin                        & 93.16                 & 92.93                  & 93.04                  & 92.59                  \\
\quad PPA  & 94.58\tiny{(+1.42)}&     95.18\tiny{(+2.25)}   &  95.49\tiny{(+2.45)}     & 94.47\tiny{(+1.88)}    \\
\quad BP \cite{roshtkhari2008novel}  & 93.68\tiny{(+0.52)}&     93.45\tiny{(+0.52)}   &  92.18\tiny{(-0.86)}     & 92.50\tiny{(-0.09)}    \\
\quad MA \cite{jahani2012automatic} & 94.39\tiny{(+1.23)}& 94.94\tiny{(+2.01)}& 94.31\tiny{(+1.27)} & 94.31\tiny{(+1.72)}           \\

\quad \cite{roshtkhari2008novel}+MC-VAE  & 94.23\tiny{(+1.07)}&     93.55\tiny{(+0.51)}   &  93.96\tiny{(+0.86)}     & 93.90\tiny{(+1.31)}    \\
\quad \cite{jahani2012automatic}+MC-VAE & 96.13\tiny{(+2.97)}& 95.19\tiny{(+2.26)}& 95.65\tiny{(+2.61)} & 94.91\tiny{(+2.32)}           \\

\quad ChrSNet \cite{zheng2022chrsnet} & 92.98\tiny{(-0.18)}  & 93.24\tiny{(+0.31)}           & 93.58\tiny{(+0.54)}           & 93.14\tiny{(+0.55)}          \\

\quad  \cite{zheng2022chrsnet}+PPA & 94.34\tiny{(+1.38)}          & 95.07\tiny{(+2.14)}          & 95.16\tiny{(+2.12)}           & 94.29\tiny{(+1.70)}          \\
\quad  MC-VAE  & 95.37\tiny{(+2.21)}         & 95.68\tiny{(+2.75)}          & 95.73\tiny{(+2.69)}         & 95.33\tiny{(+2.74)}          \\

\quad Ours & \textbf{97.56\tiny{(+4.40)}} & \textbf{97.31\tiny{(+4.38)}} & \textbf{97.46\tiny{(+4.42)}}  & \textbf{97.25\tiny{(+4.66)}} \\
ResNet-18                            &                   &                    &                 &                   \\
\quad Origin                        & 93.30                 & 92.75                  & 92.57                  & 92.62                  \\
\quad PPA  & 94.71\tiny{(+1.41)}&     94.34\tiny{(+1.59)}   &  93.89\tiny{(+1.32)}     & 93.77\tiny{(+1.15)}    \\
\quad BP \cite{roshtkhari2008novel}       & 93.72\tiny{(+0.42)} & 93.32\tiny{(+0.57)} & 93.15\tiny{(+0.58)} & 93.01\tiny{(+0.39)}   \\
\quad MA \cite{jahani2012automatic}       & 94.51\tiny{(+1.21)} & 94.76\tiny{(+2.07)} & 93.58\tiny{(+1.01)}  & 93.81\tiny{(+1.19)}     \\

\quad \cite{roshtkhari2008novel}+MC-VAE  & 94.61\tiny{(+1.31)}&     94.07\tiny{(+1.32)}   &  94.16\tiny{(+1.59)}     & 94.09\tiny{(+1.47)}    \\
\quad \cite{jahani2012automatic}+MC-VAE & 95.48\tiny{(+2.18)} & 95.29\tiny{(+2.54)} & 94.95\tiny{(+2.38)}  & 94.93\tiny{(+2.31)}           \\

\quad ChrSNet \cite{zheng2022chrsnet}     & 93.03\tiny{(-0.27)} & 93.44\tiny{\tiny{(+0.69)}}           & 93.28\tiny{(+0.71)}           & 93.34\tiny{(+0.72)}           \\

\quad  \cite{zheng2022chrsnet}+PPA & 94.57\tiny{(+1.27)}          & 94.92\tiny{(+2.17)}          & 95.30\tiny{(+2.63)}           & 94.58\tiny{(+1.96)}          \\
\quad  MC-VAE  & 95.19\tiny{(+1.89)}         & 95.81\tiny{(+3.06)}          & 95.57\tiny{(+3.00)}         & 94.86\tiny{(+2.24)}          \\

\quad Ours & \textbf{97.58\tiny{(+4.28)}} & \textbf{96.90\tiny{(+4.15)}} & \textbf{96.89\tiny{(+4.32)}} & \textbf{97.37\tiny{(+4.75)}} \\
AlexNet                              &                   &                    &                 &                   \\
\quad Origin                     & 88.19                 & 88.66                  & 87.65                  & 86.66                  \\
\quad PPA  & 90.85\tiny{(+2.66)}&     91.19\tiny{(+2.53)}   &  90.72\tiny{(+3.07)}     & 89.88\tiny{(+3.22)}    \\
\quad BP \cite{roshtkhari2008novel}       & 88.11\tiny{(-0.08)}          & 88.80\tiny{(+0.14)} & 87.50\tiny{(-0.15)}& 86.93\tiny{(+0.27)}  \\
\quad MA \cite{jahani2012automatic}       & 90.53\tiny{(+2.34)}& 91.09\tiny{(+2.43)} & 90.61\tiny{(+2.96)} & 89.77\tiny{(+3.11)} \\

\quad \cite{roshtkhari2008novel}+MC-VAE& 90.24\tiny{(+2.05)}&     90.49\tiny{(+1.83)}   &  90.12\tiny{(+2.47)}     & 89.98\tiny{(+3.32)}    \\
\quad \cite{jahani2012automatic}+MC-VAE & 91.03\tiny{(+2.84)} & 90.05\tiny{(+1.39)} & 91.07\tiny{(+3.42)}  & 90.06\tiny{(+2.31)}  \\  

\quad ChrSNet \cite{zheng2022chrsnet}     & 88.23\tiny{(+0.04)} & 88.25\tiny{(-0.41)} & 89.28\tiny{(+1.63)}& 88.92\tiny{(+2.26)}  \\

\quad  \cite{zheng2022chrsnet}+PPA & 90.36\tiny{(+2.17)}          & 90.90\tiny{(+2.24)}          & 90.10\tiny{(+2.45)}           & 89.83\tiny{(+3.17)}          \\
\quad  MC-VAE  & 91.14\tiny{(+2.95)}         & 91.21\tiny{(+2.55)}          & 90.47\tiny{(+2.82)}         & 90.29\tiny{(+3.63)}          \\

\quad Ours & \textbf{92.15\tiny{(+3.96)}} & \textbf{92.19\tiny{(+3.53)}}  & \textbf{91.89\tiny{(+4.24)}} & \textbf{91.77\tiny{(+5.11)}} \\ \hline
\end{tabular}
\end{table}

The prune ratios, masking ratios, and patch size are significant hyperparameters in our proposed approach. We begin by conducting an experiment to investigate the impact of prune ratios on the performance of our preliminary processing algorithm. As depicted in Fig. \ref{fig5}, the algorithm achieves favorable performance on the BioImLab dataset when the prune ratio is set to 0.10. We observe that the algorithm's performance remains relatively stable despite variations in the prune ratio. Consequently, the outcomes of the preliminary processing algorithm are not expected to significantly affect the performance of MC-VAE.

We then proceed to evaluate the effect of masking ratios on the performance of MC-VAE using the BioImLab dataset. Fig. \ref{fig6} presents detailed results obtained by varying the masking ratios. Across both synthetic and real-world data, we observe that the L, S, and LPIPS scores consistently increase when the masking ratio is below $70\%$, while they decrease for higher masking ratios. On the contrary, the DP and Sobel scores exhibit an inverse trend, reaching their minimum values at a masking ratio of $70\%$. These findings align with those reported in masked autoencoders \cite{he2022masked}, indicating that a higher masking ratio effectively reduces image redundancy. Consequently, it creates a reconstruction task that enables the encoder to extract as many useful features as possible, leading to a lower reconstruction error. In our scenario, the redundancy can refer to the patches within a chromosome image, excluding its density profile and key areas. The density profile is important information because it records pixel values along the medial axis of a chromosome and associates with chromosome types. Through the density profile, banding patterns and chromosome length can be roughly estimated. Regarding the key areas, they are normally specific to each chromosome. In the reconstruction task we create for MC-VAE to effectively learn image features, MC-VAE automatically identifies the key areas in chromosomes using a masking ratio setting. When the masking ratio is optimal, MC-VAE achieves the best performance by removing as much redundant information as possible.


Except for masking ratios, we also explore the effect of patch size on the performance of our proposed method. In Fig. \ref{fig6}, our approach shows superior results on both synthetic and real-world data when the patch size is set to 8x16. However, we observe a decline in performance as the patch size increases beyond 8x16, reaching its lowest point at 64x16. Larger patches introduce a higher-dimensional feature space, posing challenges for model training. Conversely, when the patch size is too small, the approach's performance also diminishes. This decline may be attributed to the limited information provided by small patches from the neighborhood, hindering the model's ability to learn long-range relationships between patches.

\begin{table}[!tb]
\centering
\caption{Performance of deep learning models for chromosome classification using original and straightened chromosomes produced by multiple methods on the Pki dataset. }
\label{t5}
\begin{tabular}{lllll}
\hline
\textbf{Model}                       & \textbf{Accuracy} & \textbf{Precion} & \textbf{Recall} & \textbf{$F_1$ Score}\\ \hline
VGG-16                               &                   &                    &                 &                   \\
\quad Origin                        & 91.82                          & 91.93                          & 91.50                         & 91.59             \\
\quad PPA         & 92.25\tiny{(+0.43)}          & 92.37\tiny{(+0.44)} & 92.26\tiny{(+0.76)} & 92.03\tiny{(+0.44)}\\
\quad  BP \cite{roshtkhari2008novel}       & 91.15\tiny{(-0.67)}          & 89.59\tiny{(-2.34) }         & 89.13\tiny{(-2.37)   }       & 89.23\tiny{(-2.36)} \\
\quad MA \cite{jahani2012automatic}       & 92.87\tiny{(+1.05)}          & 93.02\tiny{(+1.08)} & 92.50\tiny{(+1.00)}          & 92.62\tiny{(+1.03) }         \\

\quad \cite{roshtkhari2008novel}+MC-VAE  & 91.88\tiny{(+0.06)}&     90.42\tiny{(-1.51)}   &  89.95\tiny{(-0.55)}     &        89.89 \tiny{(-1.70)}    \\
\quad \cite{jahani2012automatic}+MC-VAE & 93.46\tiny{(+2.18)} & 93.65\tiny{(+2.54)} & 92.94\tiny{(+2.38)}  & 92.80\tiny{(+2.31)}    \\

\quad ChrSNet \cite{zheng2022chrsnet}     & 91.96\tiny{(+0.14)}  & 91.85\tiny{(-0.08)}          & 92.13\tiny{(+0.63)}         & 91.85\tiny{(+0.26)}          \\
\quad \cite{zheng2022chrsnet}+PPA  & 93.40\tiny{(+1.58)}          & 92.60\tiny{(+0.67)} & 92.56\tiny{(+1.06)}          & 92.09\tiny{(+0.50) }         \\
\quad MC-VAE     & 94.01\tiny{(+2.19)}  & 93.75\tiny{(+1.82)}          & 93.68\tiny{(+2.18)}         & 93.61\tiny{(+2.02)}          \\
\quad Ours                                 & \textbf{95.32\tiny{(+3.50)}}& \textbf{94.98\tiny{(+3.05)}} & \textbf{94.79\tiny{(+3.29)}}& \textbf{95.25\tiny{(+3.66)}} \\
ResNet-18                            &                   &                    &                 &                   \\
\quad Origin                   & 91.80                & 92.03                 & 91.41                 & 91.48                \\
\quad PPA                         & 92.52\tiny{(+0.72)}                 & 92.86\tiny{(+0.83)}       &           92.27\tiny{(+0.86)}           &       92.25\tiny{(+0.77)}  \\
\quad  BP \cite{roshtkhari2008novel}   & 92.39\tiny{(+0.59)}         & 92.56\tiny{(+0.53)}          & 92.83\tiny{(+1.42)}          & 92.63\tiny{(+1.15)}          \\
\quad  MA \cite{jahani2012automatic}       & 93.07\tiny{(+1.27)}          & 93.52\tiny{(+1.49)}          & 93.31\tiny{(+1.90)}          & 92.81\tiny{(+1.33)}          \\

\quad \cite{roshtkhari2008novel}+MC-VAE  & 92.63\tiny{(+0.83)}&     93.10\tiny{(+1.07)}   &  93.12\tiny{(+1.71)}     &        92.86 \tiny{(+1.38)}    \\
\quad \cite{jahani2012automatic}+MC-VAE & 94.11\tiny{(+2.31)} & 93.29\tiny{(+1.26)} & 93.46\tiny{(+2.05)}  & 93.27\tiny{(+1.79)}    \\

\quad  ChrSNet \cite{zheng2022chrsnet}     & 92.33\tiny{(+0.53)}         & 92.75\tiny{(+0.72)}          & 93.01\tiny{(+1.60)}         & 92.55\tiny{(+1.07)}          \\
\quad  \cite{zheng2022chrsnet}+PPA & 93.63\tiny{(+1.93)}          & 93.32\tiny{(+1.29)}          & 93.85\tiny{(+2.44)}          & 93.06\tiny{(+1.58)}          \\
\quad  MC-VAE  & 94.23\tiny{(+2.43)}         & 94.15\tiny{(+2.12)}          & 94.03\tiny{(+2.62)}         & 94.01\tiny{(+2.53)}          \\
\quad Ours                                 & \textbf{95.91\tiny{(+4.11)}} & \textbf{95.49\tiny{(+3.46)}} & \textbf{95.40\tiny{(+3.99)}} & \textbf{94.90\tiny{(+3.42)}} \\
AlexNet                              &                   &                    &                 &                   \\
\quad Origin                     & 89.08                 & 88.73                 & 88.65                 & 87.90                 \\
\quad PPA         &  90.71\tiny{(+1.83)}          & 89.99\tiny{(+1.26)} & 90.82\tiny{(+2.17)}          & 90.31\tiny{(+2.41) } \\
\quad  BP \cite{roshtkhari2008novel}       & 88.48\tiny{(-0.60)}         & 88.96\tiny{(+0.23)}         & 89.50\tiny{(+0.85)}          & 88.60\tiny{(+0.70) }        \\
\quad  MA \cite{jahani2012automatic}       & 91.17\tiny{+2.09)}          & 90.68\tiny{(+1.95)}          & 91.21\tiny{(+2.56)}         &  90.47\tiny{(+2.57)}          \\

\quad \cite{roshtkhari2008novel}+MC-VAE  & 89.13\tiny{(+0.05)}&     90.28\tiny{(+1.55)}   &  89.54\tiny{(+0.89)}     &        89.11 \tiny{(+1.21)}    \\
\quad \cite{jahani2012automatic}+MC-VAE & 92.06\tiny{(+2.98)} & 91.37\tiny{(+2.64)} & 92.14\tiny{(+3.49)}  & 91.25\tiny{(+3.35)}    \\

\quad  ChrSNet \cite{zheng2022chrsnet}     & 88.93\tiny{(-0.15)}          & 89.13\tiny{(+0.40)}         & 88.71\tiny{(+0.06) }        & 88.75\tiny{(+0.85)}         \\
\quad  \cite{zheng2022chrsnet}+PPA & 90.85\tiny{(+1.77)}          & 91.22\tiny{(+2.49)}          & 91.43\tiny{(+2.78)}          & 90.76\tiny{(+2.86)}          \\
\quad  MC-VAE  & 92.11\tiny{(+3.03)}         & 92.07\tiny{(+3.34)}          & 91.91\tiny{(+3.26)}         & 91.92\tiny{(+4.02)}          \\
\quad Ours                                 & \textbf{93.40\tiny{(+4.32)}} & \textbf{93.25\tiny{(+4.52)}} & \textbf{93.07\tiny{(+4.42)}} & \textbf{93.11\tiny{(+5.21)}} \\ \hline

\end{tabular}
\end{table}

\begin{table*}[!t]
\centering
\caption{External validation of methods for chromosome straightening on the ChromosomeNet dataset. ChrSNet and our methods are trained only on Pki that is independent of ChromosomeNet.}
\label{t6}
\begin{tabular}{llllll}
\hline
\textbf{Method} & \textit{\textbf{L\ score}}$\uparrow$ & \textit{\textbf{MA score}} $\uparrow$ & \textit{\textbf{Sobel score}} $\downarrow$  & \textit{\textbf{LPIPS}} $\uparrow$ & \textit{\textbf{DP score}} $\downarrow$\\ \hline
Synthetic data                       &             &             &              &             &             \\
\quad BP \cite{roshtkhari2008novel}       &$ 85.77\pm2.43 $&$ 70.54\pm5.83 $ & $267.26\pm21.83$ &$ 83.86\pm4.59 $& $141.37\pm8.97 $ \\
\quad MA \cite{jahani2012automatic}       &$ 91.93\pm3.24 $&$ 92.91\pm2.77 $ & $161.03\pm8.09 $ &$ 88.91\pm2.50 $& $132.44\pm6.22 $ \\
\quad PPA     &$ 91.96\pm2.69 $&$ 93.27\pm2.15 $ & $143.89\pm8.98$ &$89.81\pm3.01 $& $122.73\pm5.74 $ \\
\quad ChrSNet \cite{zheng2022chrsnet}    &$ 88.65\pm7.69 $&$ 79.67\pm8.66 $ & $259.66\pm20.44$ &$ 83.08\pm6.25 $& $135.66\pm16.94$ \\
\quad Ours                                 &$ \textbf{92.05}\pm\textbf{1.56} $&$ \textbf{94.49}\pm\textbf{0.71} $ &$ \textbf{84.79}\pm\textbf{14.66}$  &$ \textbf{91.32}\pm\textbf{3.18} $&$ \textbf{87.72}\pm\textbf{0.92} $  \\

Real-world data                      &             &             &              &             &             \\
\quad BP \cite{roshtkhari2008novel}       &$ 82.34\pm4.22 $&$ 70.01\pm7.69 $ & $295.52\pm19.47$ &$ 82.95\pm5.64 $& $166.10\pm14.39$ \\
\quad MA \cite{jahani2012automatic}       &$ 86.65\pm2.42 $&$ 89.59\pm2.41 $ & $179.76\pm10.94$ &$ 87.53\pm1.87 $& $135.69\pm6.90 $ \\
\quad PPA        &$ 86.70\pm2.57 $&$ 90.21\pm3.07 $ & $162.01\pm9.36$ &$ 87.97\pm1.71 $& $123.15\pm5.86 $ \\
\quad ChrSNet \cite{zheng2022chrsnet}    &$ 82.55\pm8.15 $&$ 71.29\pm10.37$ & $296.14\pm16.56$ &$ 81.16\pm6.73 $& $161.56\pm15.42$ \\
\quad Ours                                &$ \textbf{86.78}\pm\textbf{2.86} $&$ \textbf{90.92}\pm\textbf{3.20} $ & $\textbf{86.60}\pm\textbf{9.23} $   &$ \textbf{88.89}\pm\textbf{4.03} $&$ \textbf{85.43}\pm\textbf{4.31} $  \\ \hline
\end{tabular}
\end{table*}

\subsection{Potential application}

\begin{figure}[!t]
\centering
\includegraphics[scale=0.42]{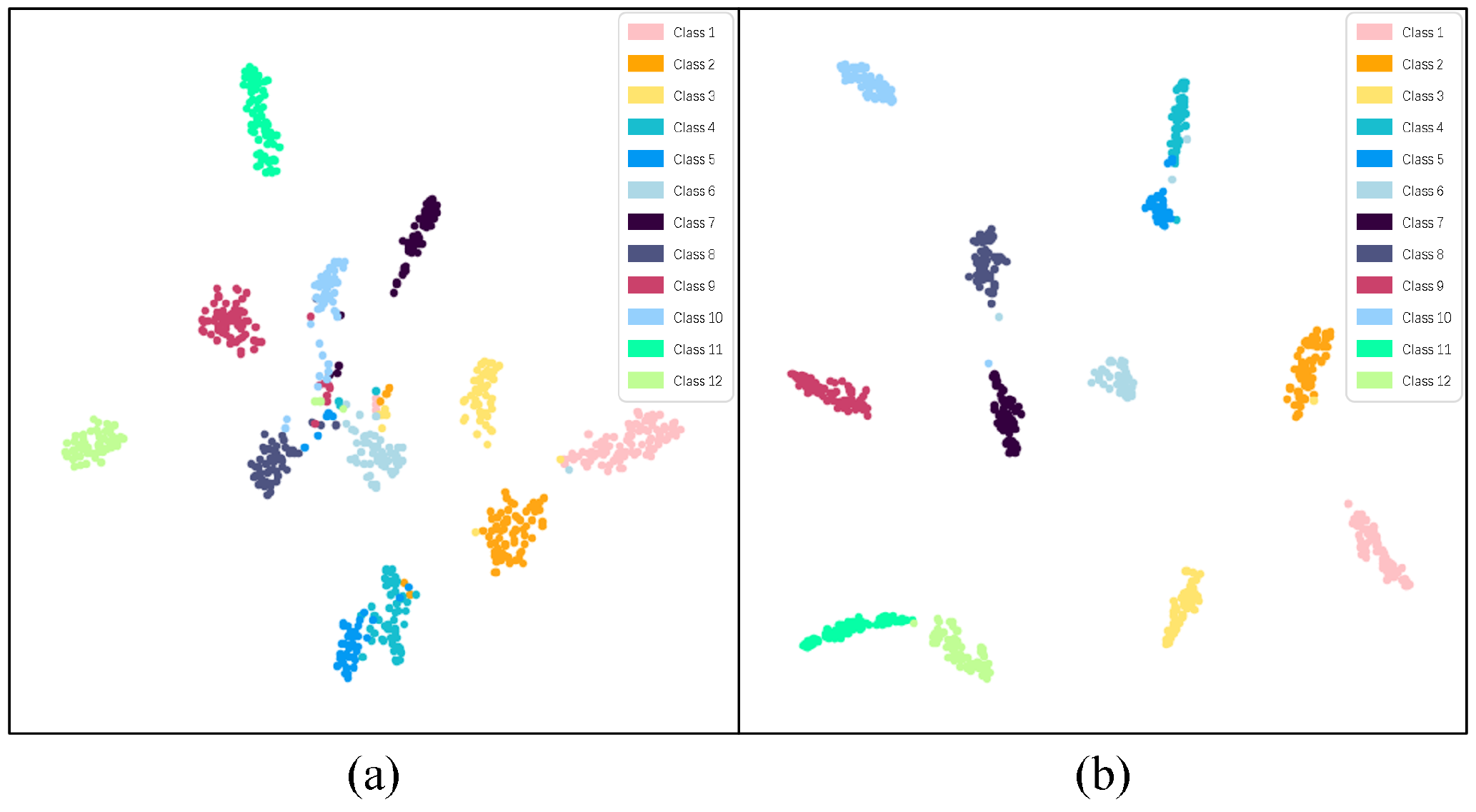}
\caption{\label{fig7} Feature projection onto a two-dimensional plane of chromosome images using the VGG classification model trained on the Pki dataset. (a) Projection results using original chromosomes. (b) Projection results using chromosomes straightened by our method. }
\end{figure}

The quality of straightened chromosomes affects the later analysis such as chromosome classification. To further verify the effectiveness of our proposed method, we perform the chromosome classification using original chromosomes and straightened ones processed by different methods. The results of three deep learning-based classification models with various chromosomes on BioImLab and Pki dataset are presented in Table \ref{t4} and \ref{t5}. Using chromosomes straightened by our method including PPA and MC-VAE achieves the best results on both datasets in terms of accuracy, precision, recall and the F1 score. Using chromosomes straightened by BP and ChrSNet shows classification results similar to that of the original data. But the classification performance related to ChrSNet can be improved by $2.17\%$ if PPA is applied before the straightening of ChrSNet on BioImLab. The MA method generates chromosomes which can enhance the accuracy of chromosome classification by $1.05\%$, $1.27\%$, and $2.09\%$ with the VGG-16, ResNet-18 and AlexNet model on the Pki dataset in Table \ref{t5}, while our method improves $3.50\%$, $4.11\%$ and $4.32\%$ on the same dataset. These findings may prove the efficiency of our method in preserving the banding patterns of chromosomes. Moreover, in addition to using MC-VAE alone, the combination of MC-VAE with BP or MA can also enhance the classification results. The consistent improvement of our framework with different deep learning models shows the potential of our method to cooperate with any system for performance enhancement of chromosome classification. To further show the advantages of using chromosomes straightened by our framework for classification, we conduct additional analysis using the VGG model on chromosomes 1-12 (easily curved) with t-distributed stochastic neighbor embedding (t-SNE). Fig \ref{fig7} depicts the t-SNE embeddings of the original chromosomes (a) and the chromosomes straightened by our framework (b). From the figure, we can observe that the intra-class distance of chromosomes is reduced, while the inter-class distance is increased. This indicates that our method enables better chromosome classification.

\begin{figure}[!t]
\centering
\includegraphics[scale=0.41]{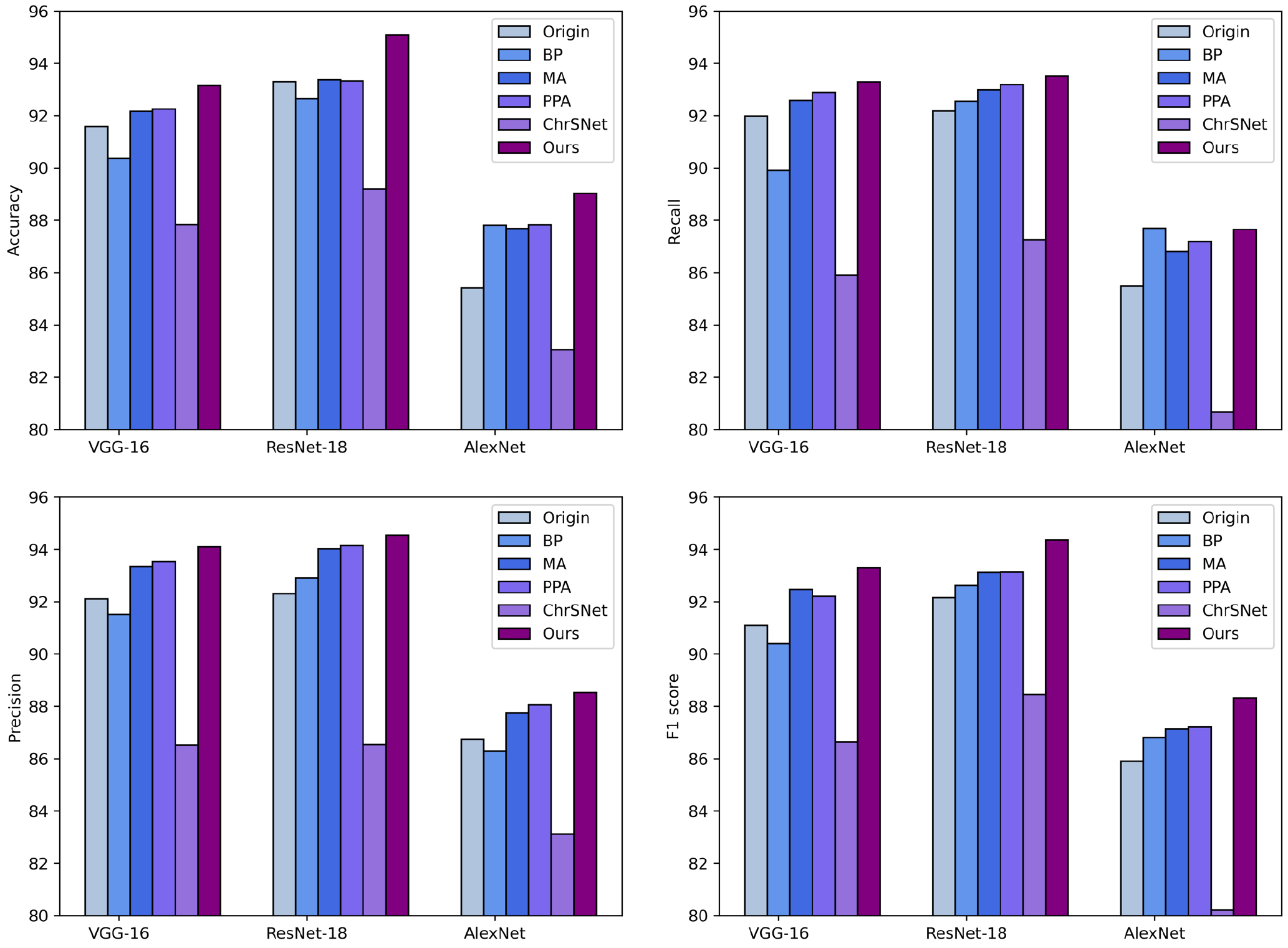}
\caption{Performance of deep learning models for chromosome classification using original and straightened chromosomes on the ChromosomeNet dataset. The straightened chromosomes are generated by the methods trained on a different dataset of Pki without seeing any data of ChromosomeNet.}
\label{fig8}
\end{figure}

\subsection{Model generalizability}

We carry out experiments to validate the generalization ability of various methods mentioned in the literature \cite{jahani2012automatic,roshtkhari2008novel, zheng2022chrsnet}, using an unseen independent dataset of ChromosomeNet. Note that deep learning models including ChrSNet and our method are trained on the data of Pki without using any data of ChromosomeNet. The validation results of methods are depicted in Table \ref{t6}. Without bells and whistles, our proposed method surpasses competitors regardless of evaluation metrics. The large scores of LPIPS ($91.32$, $88.89$) and the small scores of DP ($87.72$, $85.43$) on the synthetic and real-world ChromosomeNet data in Table \ref{t6} may suggest that the framework can effectively preserve banding patterns of chromosomes even though the chromosomes are from the dataset that is not seen by the framework. In the meantime, our approach achieves a better straight structure with a much smaller Sobel score, larger L and MA scores, and relatively smaller standard deviations compared to those of other methods. These findings show the effectiveness of our model in straightening chromosome from unseen data.

To further show the generazability of our framework, we perform experiments regarding the effect of using chromosomes of ChromosomeNet straightened by the geometric methods, or the deep learning models trained on Pki, for chromosome classification. As the results shown in Fig. \ref{fig8}, the utilization of chromosomes processed by our model presents consistent improvements for deep learning models. This might demonstrate that our model is effective and robust to unseen data with well-preserved banding features for chromosome classification.

\subsection{Subjective evaluation}
To show the potential application of our methods, we invite three senior cytogeneticists to perform a subjective evaluation of chromosome classification using both the original chromosomes and the same ones straightened by our method. Five hundred original chromosomes and their straightened ones are anonymized and randomly mixed. After evaluation, we find the accuracy of chromosome classification for every cytogeneticist is improved (84.5 $\%$ vs. 89.8$\%$, 80.1$\%$ vs. 86.7$\%$, 81.8$\%$ vs. 85.8$\%$) by using straightened chromosomes, as shown in Table \ref{t7}. Besides, compared to using original chromosomes, the results between cytogeneticists become more consistent if straightened chromosomes are used. These results show the potential clinical benefits of using chromosomes straightened by our methods for cytogeneticists in clinical practice.

\begin{table}[!t]
\centering
\caption{Classification accuracy and inter-rater variability of cytogeneticists by using original and straightened chromosomes. }
\label{t7}
\begin{tabular}{ccccccc}
\hline
     & \multicolumn{3}{c}{\textbf{Accuracy}}            & \multicolumn{3}{c}{\textbf{Cohen’s kappa}}               \\
& $C_{1}$    & $C_{2}$& $C_{3}$& $K_{12}$&$K_{13}$&$K_{23}$\\ \hline
Origin & 84.50 & 80.08 & 81.83 & 0.711 & 0.694 & 0.693 \\
Ours & \textbf{89.83} & \textbf{86.67} & \textbf{85.83} & \textbf{0.762} & \textbf{0.738} & \textbf{0.753} \\ \hline
\end{tabular}
\end{table}

\section{Conclusion and future work}
In this study, we propose a framework that utilizes patch rearrangement and conditional variational autoencoders with masking for chromosome straightening. The masked conditional variational autoencoders effectively learn the mapping between banding details of chromosome patches and curvature conditions. A high masking ratio results in a challenging image reconstruction task, which enables the model to learn local and spatial image features effectively with a convolutional neural network and a Transformer encoder. Furthermore, we design a geometric method by patch rearrangement to erase low degrees of curvature of chromosomes and to generate reasonable preliminary results for masked conditional variational autoencoders. The experimental results on three public datasets with different stain styles show the effectiveness and robustness of our method in maintaining banding patterns while keeping structure details for various types of bent chromosomes. Applying such an approach to bent chromosomes significantly improves on chromosome classification, allowing deep learning models to provide accurate karyotyping for cytogeneticists.

Although our approach has promising results for chromosome straightening on different datasets, the generalizability of the approach can be further optimized so that the method can achieve better results on unseen domains. Besides, model evaluation needs to be performed on more data before a clinical application is considered.

\section*{Declaration of competing interest}
  The authors have no relevant conflict of interest to disclose.

\section*{Acknowledgment}
We thank the Research Center for Industries of the Future at Westlake University for supporting this work.

\bibliographystyle{IEEEtran}
\bibliography{Bibliography.bib}

\end{document}